\documentclass[sigplan,screen]{acmart}
\AtBeginDocument{%
  }

\copyrightyear{2026}
\acmYear{2026}
\setcopyright{cc}
\setcctype{by}
\acmConference[ASPLOS '26]{Proceedings of the 31st ACM International Conference on Architectural Support for Programming Languages and Operating Systems, Volume 2}{March 22--26, 2026}{Pittsburgh, PA, USA}
\acmBooktitle{Proceedings of the 31st ACM International Conference on Architectural Support for Programming Languages and Operating Systems, Volume 2 (ASPLOS '26), March 22--26, 2026, Pittsburgh, PA, USA}
\acmDOI{10.1145/3779212.3790189}
\acmISBN{979-8-4007-2359-9/2026/03}

\usepackage{amsmath,amsfonts}
\usepackage{algorithmic}
\usepackage{graphicx}
\usepackage{textcomp}
\usepackage{url}
\usepackage{subfig}
\usepackage{xcolor}
\usepackage{multirow}
\usepackage{lipsum}
\usepackage{caption}
\usepackage{wrapfig}
\usepackage{hyperref}

\usepackage{makecell}
\usepackage{xspace}
\usepackage{pifont}

\usepackage{tikz}
\DeclareRobustCommand{\circled}[1]{%
  \tikz[baseline=(char.base)]{
    \node[shape=circle,fill=black,inner sep=1pt] (char)
    {\color{white}\sffamily\scriptsize #1};}}

\usepackage{caption}
\usepackage{booktabs}

\usepackage{lipsum}
\usepackage{adjustbox}
\usepackage{bm}
\usepackage{soul}
\newcommand{\minisection}[1]{\vspace{0.05in}\noindent {\bf #1}}

\usepackage{lipsum}

\newcommand{\thetitle}{\name: Value Level Parallelism For Efficient LLMs}

\newcommand{\sym}[1]{\textsf{#1}}

\newcommand{\name}{\sym{Mugi}{}\xspace}

\begin{document}

\ccsdesc[500]{Hardware~Application-specific VLSI designs}
\ccsdesc[500]{Hardware~Emerging architectures}
\ccsdesc[300]{Hardware~Arithmetic and datapath circuits}
\ccsdesc[500]{Computer systems organization~Parallel architectures}

\keywords{value-level parallelism, value reuse, data reuse, computation reuse, unary computing, temporal coding, quantization, general matrix multiplication, nonlinear approximation, large language model, KV cache, group query attention}


\title{\thetitle}





\settopmatter{authorsperrow=4}

\author{Daniel Price}
\email{daniel.price@ucf.edu}
\orcid{0009-0007-0571-7413}
\affiliation{%
  \institution{\mbox{University of Central Florida}}
  \department{Department of ECE}
  \city{Orlando}
  \state{FL}
  \country{USA}
}
\author{Prabhu Vellaisamy}
\email{pvellais@andrew.cmu.edu}
\orcid{0009-0007-7750-8725}
\affiliation{%
  \institution{\mbox{Carnegie Mellon University}}
  \department{Department of ECE}
  \city{Pittsburgh}
  \state{PA}
  \country{USA}
}
\author{John P. Shen}
\email{jpshen@cmu.edu}
\orcid{0000-0002-7225-0629}
\affiliation{%
  \institution{\mbox{Carnegie Mellon University}}
  \department{Department of ECE}
  \city{Pittsburgh}
  \state{PA}
  \country{USA}
}
\author{Di Wu}
\email{di.wu@ucf.edu}
\orcid{0000-0001-9775-8026}
\affiliation{%
  \institution{\mbox{University of Central Florida}}
  \department{Department of ECE}
  \city{Orlando}
  \state{FL}
  \country{USA}
}



\begin{abstract}
Value level parallelism (VLP) has been proposed to improve the efficiency of large-batch, low-precision general matrix multiply (GEMM) between symmetric activations and weights.
In transformer based large language models (LLMs), there exist more sophisticated operations beyond activation-weight GEMM.
In this paper, we explore \textit{how VLP benefits LLMs}.
First, we generalize VLP for nonlinear approximations, outperforming existing nonlinear approximations in end-to-end LLM accuracy, performance, and efficiency.
Our VLP approximation follows a value-centric approach, where important values are assigned with greater accuracy.
Second, we optimize VLP for small-batch GEMMs with asymmetric inputs efficiently, which leverages timely LLM optimizations, including weight-only quantization, key-value (KV) cache quantization, and group query attention.
Finally, we design a new VLP architecture, \name, to encapsulate the innovations above and support full LLM workloads, while providing better performance, efficiency and sustainability.
Our experimental results show that \name can offer significant improvements on throughput and energy efficiency, up to $45\times$ and $668\times$ for nonlinear softmax operations, and $2.07\times$ and $3.11\times$ for LLMs, and also decrease operational carbon for LLM operation by $1.45\times$ and embodied carbon by $1.48\times$.
\end{abstract}

\maketitle

\section{Introduction}
\label{sec:Introduction}

Modern artificial intelligence (AI) has been betting on deep neural networks (DNNs) for over a decade~\cite{alexnet_paper}.
A great deal of software and hardware research has been dedicated to improving the efficiency of GEMM, as it accounts for over 90\% of the total runtime~\cite{yangqingjia_dissertation}.
A key insight from this body of work is that low numerical precision gives good efficiency and accuracy.
Previous research has proposed and applied narrower data formats symmetrically to the activations and weights of GEMM, e.g. BF16~\cite{bf16_paper}, DLFloat16~\cite{dlfloat16_paper}, CBFloat~\cite{cbfloat16_paper}, FP8~\cite{fp8_paper}.
For offline workloads that process \textit{large-batch}, \textit{low-precision} data, value level parallelism (VLP) can potentially improve performance and efficiency by avoiding redundant computations, as shown by a Carat design~\cite{carat_paper}, with details given in Section~\ref{sec:vlp_review}.

\begin{figure}[!t]
    \centering
    \includegraphics[width=\columnwidth]{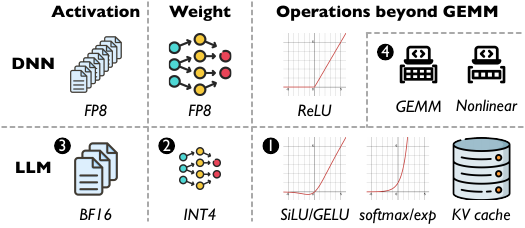}\\
    \caption{Challenges for LLM inference using VLP.}
    \label{fig:challenge_overview}
\end{figure}

\minisection{Challenges.}
Recent AI advancements have given rise to generative AI workloads, e.g., transformer-based large language models (LLMs)~\cite{attention_paper}.
LLMs exhibit complicated operations beyond activation-weight GEMM, for which VLP is designed.
Naturally, a research question is raised: \textit{can VLP address diverse LLM operations?}
Figure~\ref{fig:challenge_overview} highlights the challenges.

\circled{1} First, prior VLP architectures do not support LLM nonlinear operations, such as such as SiLU~\cite{silu_paper}, Swish~\cite{swish_paper}, GELU~\cite{gelu_paper}, and softmax.
These operations are far more complicated than the ReLU predecessor~\cite{relu_paper} and account for significant runtime if not optimized~\cite{softmax_time_1, softmax_time_2, softmax_time_3, softmax_time_4, 2021islped_uno}, despite various software~\cite{1997DivisionAlgorithms, 200964bitexpfunct} and hardware~\cite{c-lstm, exp_taylor, spatten, mobilenet, 2021islped_uno} solutions being proposed.

\circled{2} Second, prior VLP architectures misalign the trending asymmetric quantization in LLM inference, offering suboptimal efficiency.
Memory-intensive LLMs outgrow the memory capacity in mobile devices easily~\cite{llama3_paper, raspberrypi, h100_spec}.
BF16-INT4 quantization has been exploited on both weight~\cite{frantar2022gptq} and KV cache~\cite{kvquant_paper} to combat the large memory footprint while preserving accuracy.
But prior Carat only supports FP8~\cite{carat_paper}.

\circled{3} Third, prior VLP architectures are not optimized to for small-batch inputs, leading to suboptimal efficiency.
LLMs usually use a small batch size, such as 8, ensure real-time inference, since large batch sizes linearly worsen the inference latency~\cite{llm_batch_paper, llm_batch_paper2} and violate the system-level objectives (around 200~ms)~\cite{batching_paper1, batching_paper2},

\circled{4} Fourth, existing AI architectures dedicate separate matrix and vector units for  nonlinear operations and GEMM, increasing the carbon emission and lowering sustainability.
The nonlinear hardware increases on-chip area and embodied carbon during manufacture, which could outweigh the operational carbon during LLM execution, especially for more advanced technologies~\cite{act_paper}.




\minisection{Proposal.}
To overcome the challenges above, we craft \name, a new VLP architecture that support nonlinear approximation for the first time and asymmetric, small-batch GEMM, as well as reusing the array for both nonlinear operations and GEMM for efficient LLMs.
First, \name orchestrates the first-to-date VLP support for nonlinear approximation. 
\name approximates critical nonlinear operations in LLMs, such as softmax, SiLU, and GELU.
\name adopts \textit{input approximation} and generates a precise output for an approximate input, in contrast to common output approximation with precise input~\cite{c-lstm, exp_taylor, spatten, mobilenet, 2021islped_uno}.
VLP approximation is \textit{value centric} and assign greater accuracy to more important inputs.

Moreover, \name is optimized for asymmetrically quantized, small-batch GEMMs, that are not compatible in prior VLP designs~\cite{carat_paper}.
LLMs leverages weight-only quantization (WOQ)~\cite{billm_int1_paper, bitnet_int2_paper, int2_1_paper, lin2023awq, frantar2022gptq, llm_int8_paper} for activation-weight GEMM and KV cache quantization (KVQ)~\cite{zhang2024kvcache1bit,gear_paper,kvquant_paper,flexgen_paper,alisa_paper} for activation-activation GEMM, introducing BF16-INT4 GEMM.
\name supports such asymmetric quantization by \textit{customizing the data format} and \textit{optimized mapping}.
This optimization ensures high utilization for both WOQ with small-batch input and KVQ with grouped query attention (GQA) ~\cite{gqa_paper}.
We additionally \textit{minimize the buffer cost} in \name over Carat via broadcasting and output buffer leaning.

Last but not least, \name synergizes the nonlinear approximation and GEMM optimizations and maximally reuse the chip budget for LLMs. 
This allows \name to execute full LLM workloads efficiently and decrease area overhead, both of which directly correlate to a decrease in operational and embodied carbon.

The contributions of this paper are summarized as follows:
\begin{itemize}
    \item We formulate value level parallelism for nonlinear approximation, which adopts input approximation in a value-centric manner.
    \item We optimize value level parallelism asymmetric, small-batch GEMM, using timely LLM optimizations, such as quantization and group query attention.
    \item We synergize the nonlinear approximation and GEMM optimization above in one \name architecture to run full LLM workloads.
    \item We conduct experiments on multiple LLMs using \name and demonstrate good improvements in performance, efficiency and sustainability.
\end{itemize}

This paper is organized as follows.
Section~\ref{sec:Background} reviews the background.
Section~\ref{sec:Mapping} articulates VLP approximation.
Section~\ref{sec:Architecture} describes our \name architecture, with evaluations in Section~\ref{sec:Implementation} and Section~\ref{sec:Evaluation}.
Section~\ref{sec:Discussion} and Section~\ref{sec:Conclusion} discuss and conclude this paper.
\section{Background}
\label{sec:Background}

\subsection{Value Level Parallelism}
\label{sec:vlp_review}
\begin{figure}[!t]
    \centering
    \includegraphics[width=\columnwidth]{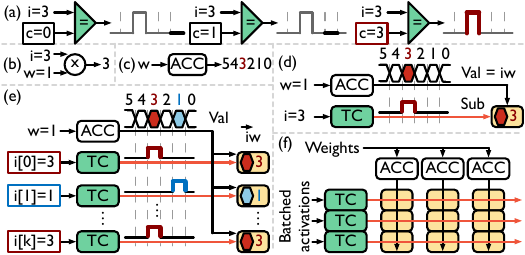}\\
    \caption{Illustration of VLP, detailed in Section~\ref{sec:vlp_review}.}
    \label{fig:vlp_review}
\end{figure}

Value level parallelism (VLP) was first proposed for GEMM operations on large-batch, low-precision data~\cite{carat_paper}, with an example for vector-scalar multiplication and vector-vector outer product given in Figure~\ref{fig:vlp_review}.
(a) shows the temporal coding of a variable $i$, done by a temporal converter (TC) in green, which is essentially an equivalence logic.
When input $i$, of value 3, equals the number in counting-up sequence (i.e., when the counter $c$ reaches 3 in the rectangle), the TC asserts a temporal spike in red at the 3rd cycle; otherwise, no spikes is generated, indicated by the bold segments in black.
(b) and (c) depicts transforming a multiplication between $i=3$ and $w=1$ into the accumulation of $w$ over time.
At cycle $i$, the accumulation outputs the correct product $iw$, equivalent to $1+1+1=3$.
(d) exemplifies \textit{temporal subscription}.
Saving the correct product (Val) into the register in yellow is enabled by the temporal spike (Sub), essentially selecting its corresponding $iw$ product of $3\times1 = 3$ in red.
(e) extends to scalar-vector multiplication between a scalar $w$ and a vector $\Vec{i}$.
The accumulation results of $w$ are shared by all vector elements.
Each vector element subscribes to its own product in parallel, based on it own input (red and blue).
We call such parallel, value-dependent sharing across multiple inputs as \textit{value reuse}.
Value reuse and temporal subscription together formulate VLP.
(f) shows that a vector-vector outer product can be obtained by organizing multiple columns of scalar-vector multiplication into a 2D array.
Carat maps batched input activations to rows and weights to columns, with the number of columns matching the temporal spike latency to avoid resource overprovision and maximize the resource utilization.
Since the temporal spike latency increases exponentially, $2^n$ cycles for an $n$-bit input, it is more beneficial to keep VLP at smaller bitwidths~\cite{2020isca_ugemm, 2021aspdac_normstab}.
Therefore, Carat opts for mapping the batch dimension to rows to achieve scalable performance with large batch sizes.

\subsection{Nonlinear Implementations}

\subsubsection{Software Implementation}
We take nonlinear operations in LLMs such as softmax, SiLU~\cite{silu_paper}, and GELU~\cite{gelu_paper} as examples, formulated in Equations~\ref{eq:softmax_shifted},~\ref{eq:silu}, and~\ref{eq:gelu}~\cite{pytorch}, where erf means error function.
To ensure numerical stability by avoiding overflow in exp, softmax inputs are usually subtracted by the maximum of all inputs.
Without KV cache, softmax can take more than $40\%$ of the total runtime in transformer models~\cite{softmax_time_1, softmax_time_2, softmax_time_3, softmax_time_4}.
The GELU function is commonly approximated as shown in Equation~\ref{eq:gelu_torch_approx} or Equation~\ref{eq:fast_gelu}~\cite{pytorch}.
The functions easily take tens even hundreds of cycles to finished~\cite{1997DivisionAlgorithms, 200964bitexpfunct}.

\noindent
\begin{center}
\begin{minipage}{0.36\linewidth}
\fontsize{7}{7}
\begin{equation}
\text{softmax} = \frac{e^{(x_i-max)}}{\sum e^{(x_i-max)}} \label{eq:softmax_shifted}
\end{equation}
\end{minipage}%
\begin{minipage}{0.26\linewidth}
\fontsize{7}{7}
\begin{equation}
\text{SiLU} \!=\!\frac{x}{1 \!+\!e^{-x}} \label{eq:silu}
\end{equation}
\end{minipage}%
\begin{minipage}{0.37\linewidth}
\fontsize{7}{7}
\begin{equation}
\text{GELU} \!=\! \frac{x}{2} \left[ 1 \!+\! \operatorname{erf}\left( \frac{x}{\sqrt{2}} \right) \right] \label{eq:gelu}
\end{equation}
\end{minipage}%
\end{center}

\noindent
\begin{center}
\begin{minipage}{\linewidth}
\fontsize{7}{7}
\begin{equation}
\text{GELU} \!=\! \frac{x}{2} \left(1 + \operatorname{Tanh}\left(\sqrt{\frac{2}{\pi}}\cdot\left(x + 0.044715\cdot x^3\right)\right)\right) \label{eq:gelu_torch_approx}
\end{equation}
\end{minipage}%
\end{center}


\begin{center}
\begin{minipage}{\linewidth}
\fontsize{7}{7}
\begin{equation}
\text{GELU} \!=\!\frac{x}{2} \left(1 + \operatorname{Tanh}\left(0.7978845608x \cdot \left(1.0 + 0.004715 \cdot x^2\right)\right)\right) \label{eq:fast_gelu}
\end{equation}
\end{minipage}%
\end{center}



\subsubsection{Piecewise Linear Hardware Approximation}
To ensure high efficiency, hardware approximations are proposed.
Piecewise linear (PWL) approximation~\cite{c-lstm, mobilenet} separates the function curve into multiple linear segments based on the input range and computes the result based on which segment an input falls into.
PWL approximations need to buffer the segment coefficients and identify the corresponding segment of an input via comparison.
For an input vector, a dedicated set of buffers, comparators, and arithmetic units is needed for each element, increasing hardware overheads.

\subsubsection{Taylor Series Hardware Approximation}
Another popular hardware approximation is a Taylor series~\cite{exp_taylor, spatten, 2021islped_uno, 2021iccd_raven, 2019islped_seco}.
The coefficient of each Taylor term is precomputed. 
With Horner’s rule, the computation can be transformed in to concatenated multiply-accumulate (MAC) operations~\cite{2021islped_uno}.
This transformation allows for vectorized implementation, where coefficients can be shared by all inputs efficiently.
However, Taylor approximation exhibits poor accuracy when inputs are far from the Taylor expansion point.
Allowing more points introduces addition hardware overheads.

\minisection{Summary.}
Different implementations offer distinct accuracy and efficiency tradeoffs, and this work introduces a novel VLP approximation for nonlinear operations.

\subsection{Large Language Model Inference}
\label{subsec:autoregressive}

Modern LLMs are mainly built on attention-based transformers~\cite{attention_paper}.
During prefilling, multiple tokens are processed in parallel, resulting in GEMM operations.
During decoding, one token is processed, resulting in GEMV operations, unless input tokens are batched.
However, even with batched input, normal attention still performs GEMV for KV cache~\cite{kv_cache_paper}.


\subsubsection{Grouped Query Attention}
GEMV in normal attention severely lowers the hardware utilization~\cite{alisa_paper}.
To mitigate the problem, grouped query attention (GQA) is proposed, where multiple Q tokens share the same KV cache~\cite{gqa_paper}, creating small-batch GEMM.
In this paper, \name benefits from GQA to improve the hardware utilization.

\subsubsection{Weight-Only Quantization}
Low-bit quantization is now the de facto technique to reduce the memory footprint~\cite{fp8_quant_paper}.
Most prior works adopt symmetric quantization, e.g., INT8 or FP8 for both inputs and weights~\cite{tensorrtllm, fp8_quant_paper, llama3_paper, liu2024deepseek}, which are still too memory inefficient for LLMs.
Developers resort to sub-byte quantization for weights, while keeping the inputs in floating point format, e.g, BF16-INT4 weight only quantization (WOQ)~\cite{billm_int1_paper, bitnet_int2_paper, int2_1_paper, lin2023awq, frantar2022gptq, llm_int8_paper}.

\subsubsection{KV Cache Quantization}
KV cache introduces additional memory footprint on top of weights, leading to potential out-of-memory errors~\cite{alisa_paper}.
BF16-INT4 KV cache quantization (KVQ) has been leveraged to compress KV cache with minimum accuracy drop~\cite{zhang2024kvcache1bit,gear_paper,kvquant_paper,flexgen_paper,alisa_paper}.
Moreover, WOQ and KVQ can be combined together, reporting just a 0.02 increase in perplexity~\cite{kvquant_paper}.

\minisection{Summary.}
These LLM optimizations are timely and allow optimizing asymmetric, small-batch GEMM efficiently.

\subsection{Sustainable Computing}
\label{subsec:sustainable_computing}
Carbon emissions have become a growing concern in the AI world~\cite{ITU2025GreeningDigitalCompanies} and AI inference reportedly contributes up to 90\% of datacenter costs~\cite{aws_ai}.
To quantify carbon emissions, carbon modeling focuses on \textit{operational} carbon for workload deployment and \textit{embodied} carbon for infrastructure manufacture over the full lifetime~\cite{wu2022sustainableaienvironmentalimplications, faiz2024llmcarbonmodelingendtoendcarbon, patterson2021carbonemissionslargeneural}.
The equivalent emissions ($\text{CO}_{2\text{eq}}$) are formulated in Equation~\ref{eq:operational_carbon}, where E, CI, CPA are short for energy and carbon intensity, carbon emitted per unit area.
Embodied carbon is taking over operational carbon as the majority of contributed emissions~\cite{wu2022sustainableaienvironmentalimplications, faiz2024llmcarbonmodelingendtoendcarbon}.

\noindent
\begin{center}
\begin{minipage}{0.47\linewidth}
\fontsize{7}{7}
\begin{equation}
\text{Operational CO}_{2\text{eq}} = E\cdot CI \label{eq:operational_carbon}
\end{equation}
\end{minipage}%
\begin{minipage}{0.51\linewidth}
\fontsize{7}{7}
\begin{equation}
\text{Embodied CO}_{2\text{eq}} = Area \cdot CPA \label{eq:embodied_carbon}
\end{equation}
\end{minipage}%
\end{center}

\minisection{Summary.} Our \name shares the hardware for both nonlinear operation and GEMM, and contributes to reduction of both operational and embodied carbon.

\section{VLP Nonlinear Approximation}
\label{sec:Mapping}

\begin{figure*}[!t]
    \centering
    \includegraphics[width=\textwidth]{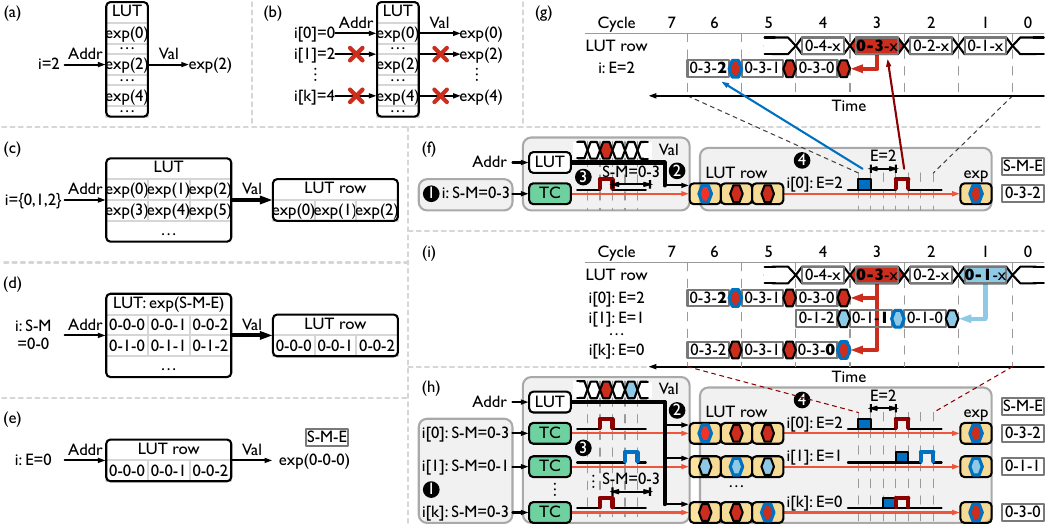}
    \Description{}
    \caption{
    VLP approximation for nonlinear operations, exp here, with a floating-point input $i$, represented as S-M-E, denoting the sign, mantissa and exponent.
    More details are in Section~\ref{subsec:formulation}.
    }
    \label{fig:vlp_for_nonlinear}
\end{figure*}

\subsection{Formulation}
\label{subsec:formulation}
We formulate VLP approximation as in Figure~\ref{fig:vlp_for_nonlinear}.
(a) depicts a conventional lookup table (LUT) for exp, indexing with an address and receiving its corresponding value.
(b) indicates that such a conventional LUT can only sequentially process different inputs, limiting the scalability.
To alleviate this restriction, (c) splits the lookup processes into two steps.
First, a row of exp values with consecutive input values are retrieved from the LUT, then the correct value can be selected from that row.
Such a split inspires VLP approximation.
(d)-(e) illustrates this split within VLP.
(d) uses an input sign and mantissa (i: S-M) to index the LUT row, and (e) then uses the exponent (i: E) to select the final result from the row.


(f) details VLP approximation within a single row.
\circled{1}--\circled{4} denotes four phases, i.e., input field split, value reuse, mantissa temporal subscription, and exponent temporal subscription.
\circled{1} the input field split phase splits the input S-M-E (0-3-2) into S-M (0-3) and E (2), as in (d)-(e).
\circled{2} the value reuse phase organizes the LUT the same as that in (d), where each LUT row contains all values for the same S-M.
At each cycle, an ascending address is sent to the LUT, and generates an output LUT row, marked by bold lines.
Overtime, LUT rows are reused by different S-M values.
\circled{3} the mantissa temporal subscription phase further splits the S-M to reuse LUT rows.
S-M (0-3) generates a temporal signal via the temporal converter in green, subscribing to the row at the 3rd cycle with a red fill.
The LUT row will be stored into the yellow blocks when a temporal spike arrives.
\circled{4} the exponent temporal subscription leverages the temporal spike of the exponent (E=2) to subscribe the final exp results from the LUT row, selecting the value at the 2nd cycle with a blue outline.
\textit{Starting from the moment} when the correct LUT row is subscribed in \circled{3}, the exponent also starts generating its own temporal spikes.
Therefore, the full VLP approximation requires the total duration of both mantissa and exponent temporal spike timing to finish.

(g) zooms into the single row for mantissa and exponent temporal subscription.
The LUT row subscription is indicated by S-M.
`-x' here denotes all exponents.
The corresponding rows will be sent to the proper inputs, indicated by the bold lines.
The exponent subscription is indicated by the blue outlined blocks.
Following (f), this example selects the row corresponding to an S-M of (0-3), subscribing at the 3rd cycle, then subscribing where E=2, or at the 2nd cycle of \circled{4} in (f). 
This full process takes a total of 6 cycles, which is the sum of two subscription.

(h-i) expands approximation to a full array, enabling approximation of vector $\Vec{i}$.
By leveraging VLP, selected rows can be shared across the array in parallel, individually subscribing to their final result.

\subsection{Input Approximation}
VLP approximation favors lower-precision inputs to reduce the duration of temporal spikes.
However, popular data formats have a wide mantissa field, e.g., BF16 mantissa has 7 bits.
Therefore, in the input field split phase, we round the input mantissa to fewer bits for softmax, SiLU and GELU.
Profiling shows that rounding introduces uniform errors to the input mantissa values, as the mantissa are uniformly distributed in softmax, SiLU and GELU, and this pattern is consistent across models and modalities.
We exclude these results for simplicity.

\begin{figure*}[!t]
    \centering
    \includegraphics[width=\textwidth]{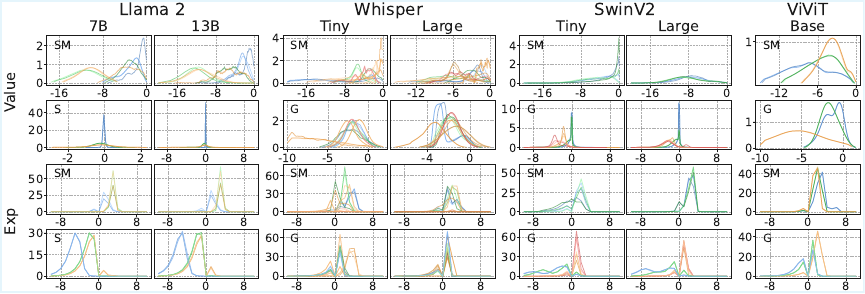}\\
    \Description{}
    \caption{
    Distribution of input values and exponents of nonlinear operations in transformer models.
    Profiled layers, stages, and sequence lengths are detailed in Table~\ref{tab:studied_llms}. 
    Cooler colors represent early layers, while warmer colors represent later layers. 
    Within each color, lighter lines represent shorter sequence lengths, while darker lines represent longer sequence lengths. 
    Softmax, SiLU, and GELU are abbreviated as SM, S, and G, denoting the nonlinear function within each window.
    }
    \label{fig:nonlinear_distribution}
\end{figure*}

\subsection{Value-Centric Approximation}
\begin{figure}[!t]
    \centering
    \includegraphics[width=0.75\columnwidth]{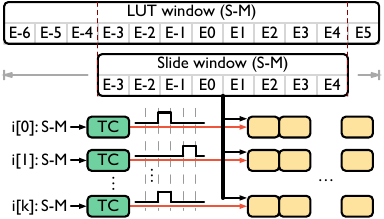}
    \Description{}
    \caption{An example sliding window for input mapping.
    This example chooses the exponent range of $[-3, 4]$ for the current set of inputs, from the full LUT window with the exponent range of $[-6, 5]$.
    The sliding window size of 8 is chosen to match the VLP array width in prior VLP works~\cite{carat_paper}. 
    This sliding window can slide left and right for each mapping, aiming to minimize the accuracy loss.
    }
    \label{fig:vlp_nonlinear_sliding}
\end{figure}

The approximation efficiency also suffers from wide exponents, since both the temporal signal length and the LUT row size grows exponentially with exponent bitwidth.
Our formulation leverages the insights that input exponents are often clustered at a small range.
We focus on these important values, following a value-centric approach.
We profile the input distribution of softmax, SiLU and GELU in Figure~\ref{fig:nonlinear_distribution}.
For softmax, exponent values are concentrated around $[-3, 4]$, despite input values being widely spread.
Similar observations also exist in SiLU and GELU.
We define these model-specific, important exponents as the LUT window, and we only store the results for these exponents in the LUT.
However, a single mapping, with a set of inputs for value reuse, might not cover the full range of important exponents.
Therefore, we opt for a sliding window for each mapping and choose an optimal range, as shown in Figure~\ref{fig:vlp_nonlinear_sliding}.

\begin{figure*}[!t]
    \centering
    \includegraphics[width=\textwidth]{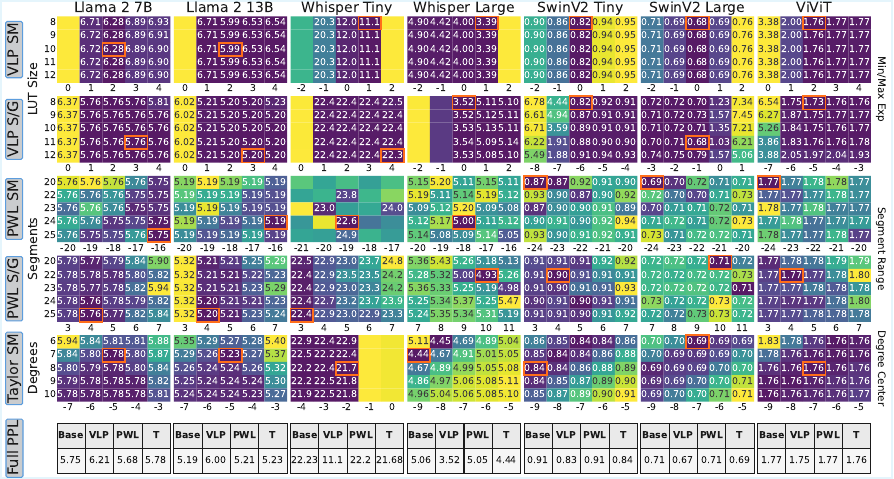}\\
    \Description{}
    \caption{
    Perplexity and loss heatmaps of transformer models, showing the best for each nonlinear operation separately highlighted in orange (lower is better).
    Llama 2 and Whisper show perplexity, while Swin and Vivit show Loss.
    Combined softmax/activation metrics are short right of the title, with VLP being on top and PWL on bottom.
    Boxes on the left denote the approximation method, with softmax, SiLU and GELU abbreviated as SM, S, and G.
    The labels next to the boxes denote y axis value and \textit{the labels on the right side denote x axis}.
    LUT size refers to the number of exponents stored in the LUT, and Min/Max exp denotes the maximum or minimum value used to create the LUT.
    For PWL, segment range (sr) denotes the approximation range, with SM going from [sr, 0] and S/G going from [-sr, sr]. For taylor series, degrees denotes the number of polynomial expansions used, while degree center is the center of the expansion.
    Empty boxes represent masked large values. Each column’s table lists full end-to-end perplexity values (SM and S/G). The Taylor-series values, denoted T, include only SM.
    }
    \label{fig:nonlinear_heatmap}
\end{figure*}

\begin{figure}[!t]
    \centering
    \includegraphics[width=\columnwidth]{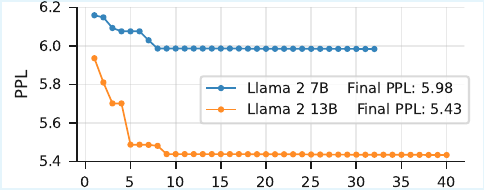}
    \caption{Per-layer perplexity tuning of Llama 2 (7B, 13B), with final achieved perplexity noted in the legend. Tuning is done progressively across layers.
    }
    \label{fig:layer_tuning}
\end{figure}

\begin{figure}[!t]
    \centering
    \includegraphics[width=\columnwidth]{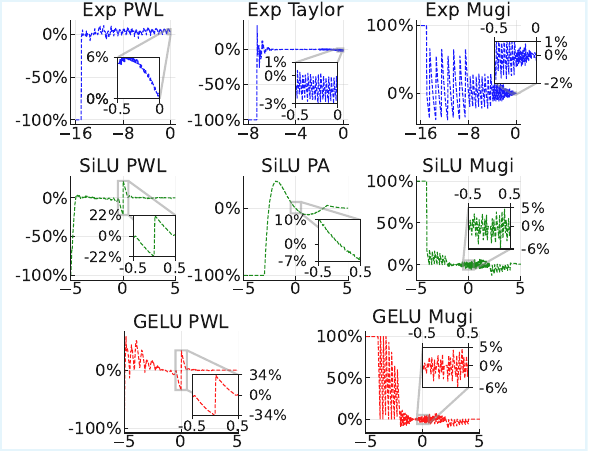}
    \Description{}
    \caption{Relative errors against software implementation. 
    The most accurate configurations from Figure~\ref{fig:nonlinear_heatmap} are compared.
    x and y axes are input values and relative error.
    $100\%$ error indicates flushing output to 0.
    }
    \label{fig:nonlinear_error}
\end{figure}

\subsection{Accuracy Impact}
\label{sec:acc_mugi}

We explore the accuracy of different window sizes and boundaries as in Figure~\ref{fig:nonlinear_heatmap}.
We compare VLP approximation against prior approximations, like Taylor series~\cite{exp_taylor, spatten}, piecewise linear (PWL) approximation~\cite{c-lstm}, and partial approximation (PA)~\cite{mobilenet}.
For most models with full VLP approximation (combined softmax/activation), \name shows better accuracy, except for Llama 2, whose softmax distribution varies significantly across layers, as shown in Figure~\ref{fig:nonlinear_distribution}.
To address this, Figure \ref{fig:layer_tuning} demonstrates per-layer tuning of Llama 2, selecting the optimal LUT range for each layer.
This mitigates accuracy loss, yielding perplexity values approaching those in line with other approximation techniques.

Figure~\ref{fig:nonlinear_error} further shows the accuracy of the nonlinear approximation.
While VLP approximation does not exhibit the best error, it has the best accuracy where inputs are important, in term of magnitude and quantity.
For softmax in layer 0, high exp accuracy for majority of the inputs (Figure~\ref{fig:nonlinear_distribution}) propagates less errors to deeper layers.
In deeper layers, more inputs center around $-10$.
Their output magnitudes are smaller, e.g, adding 22k $e^{-10}$ equals $e^{0}$.
Therefore, even though VLP approximation is less accurate, it has negligible impacts compared to inputs closer to 0.
Another contributor to overall accuracy is that uniform input errors from input approximation can cancel out each other's output errors during summation.
For SiLU/GELU, inputs consistently cluster around 0, where VLP approximation is more accurate.

\section{\name Architecture}
\label{sec:Architecture}

\begin{figure*}[!t]
    \centering
    \includegraphics[width=\textwidth]{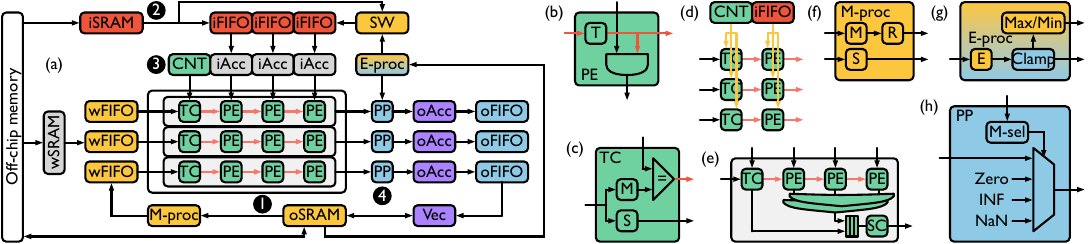}\\
    \Description{}
    \caption{
    Overview of \name architecture.
    \circled{1}--\circled{4} (yellow, red, green, and blue) correspond to \circled{1}--\circled{4} phases of VLP approximation in Figure~\ref{fig:vlp_for_nonlinear}.
    Purple is for mapping softmax, and gray marks the additional hardware for GEMM.
    }
    \label{fig:architecture_overview}
\end{figure*}

We introduce \name, a novel VLP architecture to support nonlinear approximation and asymmetric, small-batch GEMM, as well as reusing the array for both nonlinear operations and GEMM for efficient LLMs.
\name supports nonlinear softmax, SiLU and GELU operations.
\name supports BF16-INT GEMM, leveraging timely GQA, WOQ and KVQ optimizations.
The support for both nonlinear operations and GEMM above are mixed in one singular architecture with maximized resource reuse.
Figure~\ref{fig:architecture_overview} outlines our \name architecture.
It double buffers all memory hierarchies to hide access latency.
We follow prior VLP designs and set the number of columns to 8 (matching 3-bit mantissa) for optimal performance and efficiency tradeoffs~\cite{carat_paper}.

\circled{1} is for the input field split phase.
M-proc splits the sign (S) and mantissa (M) fields for BF16 input and approximates the mantissa to 3-bit via rounding (R).
For a given mapping, E-proc processes the exponent (E) values to determine the maximum or minimum exponent, which determines the LUT sliding window in the SW block.
The sliding window size is fixed to 8 to match array width.
It also clamps the exponent, underflowing to 0, and overflowing depending on the nonlinear operation. 
In softmax, overflow values are set to the maximum value of the LUT, while SiLU/GELU passes values through directly.
The exponent is sent to post processing (PP) block for the final result.

\circled{2} is for the value reuse phase.
The iSRAM acts as the LUT, and the pre-computed, output LUT window is sent to the SW block to generate the sliding window.
The sliding window is sent to iFIFO to stagger the input by one cycle to adjacent columns.
This staggering ensures fully pipelined execution in our VLP approximation.

\circled{3} is for the mantissa temporal subscription phase.
The temporal converter (TC) converts the approximate mantissa (M) to a temporal signal using the counter (CNT) and leaves the sign (S) to PP.
The temporal signal is then pipelined in a row via the T register in the processing element (PE).
Temporal subscription is done using the AND gate in each PE.
Within a PE column (Figure~\ref{fig:architecture_overview}~(d)), both the counter value and sliding window are broadcast.
Within a PE row (Figure~\ref{fig:architecture_overview}~(e)), the subscribed results will be sent out via OR gates, since only one column will be activated by the pipelined temporal spike.
Two sets of OR gates, together with a small FIFO, double buffer the results from two spikes.
Sign conversion (SC) XORs all signs to generate the final result.

\circled{4} is for the exponent temporal subscription phase.
The PP block takes the exponent from the E-proc and generates a MUX selection signal.
If no special values exist, this selection signal is the temporal spike from the exponent, subscribing the correct element in the sliding window.
If there are special values, the multiplexer outputs the proper special values among Zero, infinity (INF) and Not-a-Number (NaN).

\subsection{Nonlinear Approximation}
\label{subsec:proposed_for_nonlinear}

\begin{figure*}[!t]
    \centering
    \includegraphics[width=\textwidth]{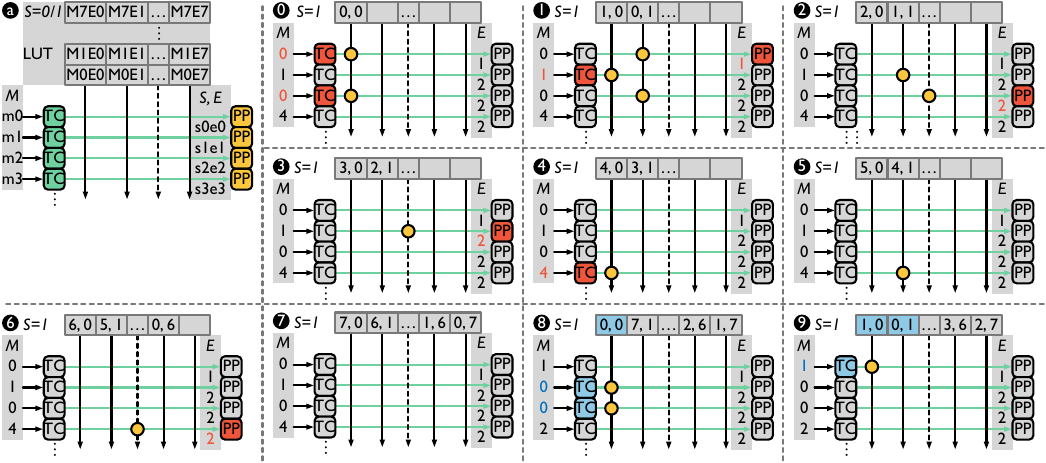}\\
    \Description{}
    \caption{Mapping element-wise nonlinear operations to \name.
    \circled{a} abstracts the VLP array for nonlinear operations.
    M, S and E are for mantissa, sign and exponent.
    \circled{0}--\circled{9} are examples for exp, and the numbers denotes the clock cycles.}
    \label{fig:nonlinear_mapping}
\end{figure*}

To better understand how \name works, we give a walkthrough example in Figure~\ref{fig:nonlinear_mapping}.
Here, rows apply broadcasting, while columns adopt pipelining.
The LUT (iSRAM) stores pre-computed nonlinear results, and each LUT row contains a vector of results for one mantissa.
The LUT size will double if the nonlinear operation has both positive and negative inputs.
In \circled{0}--\circled{7}, TC or PP is red if a temporal signal (yellow) arrives.
In \circled{8}--\circled{9},
a new mapping is marked with blue.

In the first input field split phase, the 8-bit BF16 mantissa is rounded to INT4 with a 3-bit mantissa magnitude to generate an 8-cycle temporal signal.
In the second value reuse phase, LUT row vectors are read out per cycle in a mantissa-ascending order, and reused in the next subscription phase.
Note that vectors from different LUT columns are sent to the array in a staggered manner, as the temporal signal is pipelined across columns.
In the third mantissa temporal subscription phase, temporal signals are generated from approximated mantissa.
For each mantissa, the TC turns red upon the coincidence of the input value and equivalent clock cycle.
The fourth exponent temporal subscription phase subscribes to the correct result from the LUT vector, indicated by the red exponent (e).
The sign (s) is omitted here as it is always negative. 
The cycle index to get the final result is the sum of mantissa and exponent values.
After 8 cycles of red input, at cycle 8, new blue inputs enter the array.

The above VLP approximation works for element-wise nonlinear operations, e.g., exp and SiLU/GELU.
Additional summation and division are needed for softmax.
We first compute the exp for all inputs (maximum subtracted).
To perform the summation, when we compute exp, the output accumulator (oAcc) simultaneously accumulates the exp results.
Once all exp operations finish, we store the sum to the oSRAM from the oFifo.
Next, we divide all exp results by the sum using the vector multiplication array (Vec) in Figure~\ref{fig:architecture_overview}. 
This array multiplies the exp by the reciprocal of the sum in one cycle.
To ensure high utilization, we map both attention head and batch across rows for softmax.

\subsection{GEMM Optimization}


\name optimizes GEMM over prior VLP designs in two ways.

\minisection{Format Customization.}
Asymmetric BF16-INT4 GEMM imposes challenges to Carat.
As Carat maps FP8 input across rows, changing input to BF16 will prolong the temporal signal from 8 (3-bit mantissa magnitude) to 128 cycles (7-bit), prohibitively lowering the throughput.
If INT4 weights were mapped to rows, the FP8 datapath is wasted.
To overcome these, \name transposes the mapping, i.e., INT4 weight/KV cache to rows with a slim INT4 datapath, and BF16 input/Q token to columns.
This mapping offers high utilization, since LLM tokens are large enough to fill in all rows, and 8 Q tokens in a group for GQA can fill in all columns.
The customization timely synergizes with the trends of LLMs, small batch sizes, large token sizes, WOQ, KVQ and GQA, none of which are compatible in Carat. 
WOQ and KVQ require dequantization after GEMM, which is done by the vector array.

\minisection{Buffer Minimization.}
Buffers (FIFOs) occupy significant area in Carat, due to pipelining the inputs across rows and double buffering in the output OR tree.
The relevant cost scales quadratically with the array size.
\name solves this problem via broadcasting and output buffer leaning (optimizing two FIFOs into one without functional changes), successfully lowering the total buffer area by $4.5\times$.

With the optimizations above, GEMM can be exectued efficiently, following the flow in Figure~\ref{fig:vlp_review}.
To ensure scalability, we can further use a 2D mesh Network-on-Chip (NoC) to connect multiple nodes.
We consider output stationary dataflow and inter-node accumulation, and GEMMs are evenly tiled across nodes to enhance efficiency and utilization.

\section{Experimental Setup}
\label{sec:Implementation}

\subsection{Large Language Model}
\label{subsec:large_language_model_setup}

The evaluated LLMs are summarized in Table~\ref{tab:exp_llm}, with LLaMA2-70B supporting GQA using a group size of 8. 
We implement all models using the HuggingFace Transformers Package~\cite{hugging_face_transformers}, profiling and computing loss or perplexity for each model over 100 inferences. During profiling, we extract the runtime input nonlinear tensors across all tokens and record the value and exponent distributions, which is documented in Figure~\ref{fig:nonlinear_distribution}. To obtain the perplexity and loss values shown in Figure~\ref{fig:nonlinear_heatmap}, we sweep various configurations for each nonlinear implementation, and show windows containing the best performing configuration. For both figures, we show the smallest and largest model of each model family, with the only exception of Llama 2 13B replacing Llama 2 70B due to memory and runtime limitations.

\begin{table}[!t]
    \centering
    \caption{Studied LLMs in this paper.}
    \label{tab:exp_llm}
    \begin{adjustbox}{max width=\linewidth}
    \begin{tabular}{l|c|c|c|c|c|c|c|c}
        \toprule
        \multirow{2}{*}{\textbf{Model}} & \multicolumn{3}{c|}{\textbf{Llama2~\cite{llama2_paper}}} & \multicolumn{2}{c|}{\textbf{Whisper~\cite{whisper_paper}}} & \multicolumn{2}{c|}{\textbf{SwinV2}} & \textbf{ViViT}\\
    \cmidrule{2-9}
         & \textbf{7B} & \textbf{13B} & \textbf{70B} & \textbf{tiny} & \textbf{large} & \textbf{tiny} & \textbf{large} & \textbf{base}\\
        \midrule
        \midrule
        Batch size &  \multicolumn{7}{c}{1 - 32} \\
    \cmidrule{2-9}
        \# layers & 32 & 40 & 80 & 4 & 32 & 12 & 24 & 12 \\
        \# stages & & & & & &  4 & 4 &\\
        \# attn heads & 32 & 40 & 64 & 6 & 20 & 3 - 24 & 6 - 48 & 12\\
        \# KV heads & 32 & 40 & 8 & 6 & 20 & 3 - 24 & 6 - 48 & 12\\
        Attn h dim  & 4096 & 5120 & 8192 & 384 & 1280 & 96-768 & 192-1536 & 768\\
        FFN h dim & 11008 & 13824 & 28672 & 1536 & 5120 & 384-3072 & 768-6144 & 3072\\
    \cmidrule{2-9}
        Seq len & \multicolumn{3}{c|}{4096} & \multicolumn{2}{c|}{1500} & \multicolumn{2}{c|}{64-4096} & 3136\\
    \midrule
    \midrule
    Prof. layers & 1/16/32 &  1/20/40 &  1/32/64 & 1/3/6 &  1/10/20 & 1/12 &  1/24 & 1/6/12 \\
    \cmidrule{2-9}
    
          
        \multirow{3}{*}{Prof. seq len} 
          & \multicolumn{3}{c|}{1024}
          & \multicolumn{2}{c|}{112/224} 
          & \multicolumn{2}{c|}{16/32} 
          & 784 \\

          & \multicolumn{3}{c|}{2048} & \multicolumn{2}{c|}{375/750} 
          & \multicolumn{2}{c|}{64/1024} 
          & 1568 \\

          & \multicolumn{3}{c|}{4096} & \multicolumn{2}{c|}{1500} & \multicolumn{2}{c|}{2048/4096} & 3136 \\
        \bottomrule
        \multicolumn{9}{l}{\scriptsize{* $h =$ hidden, Prof. = Profiled}}
    \end{tabular}
    \end{adjustbox}
    \label{tab:studied_llms}
\end{table}


\begin{table}[!t]
    \centering
    \caption{Comparison of \name and baselines: SA refers to systolic array, SD to SIMD array, Tensor to Tensor Core, with off-chip bandwidth at 256 GB/s. Input (i), weight (w), and output (o) refer to respective components, with ranges (e.g., a-b) covering all powers of 2 between a and b. Input word config applies to the query word, and weight word to key and value words with kvcache quantization. -S denotes scaled-up configurations. All designs use 4x4 and 8x8 NoC layouts except Tensor and -S configurations, which use a 2x1 and 2x2 and no NoC, respectively.}
    \begin{adjustbox}{max width=\linewidth}
    \begin{tabular}{c|c|c|c|c|c|c|c}
        \toprule
        \multirow{1}{*}{\textbf{Configuration}} & \textbf{\name} & \textbf{Carat} & \textbf{SA} & \textbf{SD} & \textbf{SA-S} & \textbf{SD-S} & \textbf{Tensor}\\
        \midrule
        \midrule
        i/w/o SRAM & \multicolumn{4}{c}{64KB} & \multicolumn{3}{|c}{1MB} \\
        \cline{2-8}
        Array height (H) & \multicolumn{2}{c}{32 - 256} & \multicolumn{2}{|c}{4 - 16} & \multicolumn{2}{|c|}{32 - 64} & \multicolumn{1}{c}{16} \\
        \cline{4-7}
        Array width (W) & \multicolumn{2}{c}{8} & \multicolumn{4}{|c|}{H}  &\multicolumn{1}{c}{8} \\
        \cline{2-8}
        Array Depth (D) & \multicolumn{6}{c|}{N/A} & \multicolumn{1}{c}{16} \\
        \cline{2-8}
        \multirow{1}{*}{Input word} & \multicolumn{7}{c}{16} \\
        \multirow{1}{*}{Weight word} & \multicolumn{7}{c}{4} \\
        \cline{2-8}
        \multirow{1}{*}{NoC shape} & \multicolumn{4}{c}{4x4, 8x8} & \multicolumn{2}{|c|}{N/A} & \multicolumn{1}{c}{2x1, 2x2}\\
        \bottomrule
    \end{tabular}
    \end{adjustbox}
    \label{tab:hardware config}
\end{table}


\subsection{Hardware}
\subsubsection{\name}

As shown in Table~\ref{tab:hardware config}, we set the oSRAM width to enable wFIFO loading of nonlinear operations in 8 cycles, ensuring sufficient bandwidth. 
The wSRAM width is similarly configured to allow loading in 8 cycles for GEMM operations. \name's vector array is configured to scale array outputs after exiting the oFIFO, hiding latency.
\name is optimized for output stationary outer-product computation.

\subsubsection{Baseline}
\label{subsubsec:baseline_setup}
We build baselines with components for both nonlinear operations and GEMM.

Nonlinear approximation hardware uses alternative vector arrays with added components to implement the Taylor series and PWL approximation methods~\cite{exp_taylor, c-lstm}.
The Taylor series is implemented with Horner's method up to 9 degrees, and requires additional registers to store coefficients. 
PWL implementation adopts 22 segments, requiring additional registers and comparators to store and select proper segments.
Both methods are configured to achieve their best perplexity as shown in Figure~\ref{fig:nonlinear_heatmap}.
We also consider a vector array of MAC units to precisely compute nonlinear operations, which require 44 cycles~\cite{1997DivisionAlgorithms, 200964bitexpfunct}.


For GEMM, we compare \name with Carat~\cite{carat_paper}, systolic arrays, SIMD arrays, FIGNA~\cite{figna_paper}, and tensor cores from Nvidia Hopper GPUs~\cite{h100_spec}, as well as a \name-L design.
Given Carat only supports FP8 GEMM with inputs mapped across rows, we modify its accumulators at the top to BF16 and map inputs across columns, while using its FP8 data path for INT4 weights.
The systolic and SIMD arrays are similar, with the systolic array needing additional control hardware and a column of output accumulators, compared to SIMD's adder trees.
Like \name, Carat implements output stationary, while the systolic and SIMD arrays use weight stationary configurations.
The FIGNA configurations extend both systolic and SIMD arrays with the FIGNA PE, which is customized for FP-INT multiplication.
Additionally, scale-up versions of both systolic and SIMD arrays with MAC and FIGNA configurations are evaluated. 
The tensor core has GEMM shaped as 8x16x16, and is a fully pipelined design to perform 8x16x16 MAC operations per cycle.
We only compare tensor core in NoC settings.
The \name-L uses a dedicated LUT to approximate nonlinear operations, rather than temporal coding based approximation.
We ensure 8 inputs share one LUT to match the throughput of \name.
Similar to \name, all design's wSRAM and oSRAM widths are selected to load the array without introducing additional latency.

\subsubsection{Network on Chip and Off-Chip Memory}
The NoC incorporates three channels for input, weight, and output. 
An output-stationary approach is employed across all implementations. 
Each design operates with a NoC frequency of 400 MHz and a HBM bandwidth of 256 GB/s from off-chip memory.
Both the NoC and off-chip memory are configured such that they always supply the minimum bandwidth required to not bottleneck computation.

\subsection{Carbon Modeling}
To model both \name and baselines operational and embodied carbon, we follow a consistent approach by previous works~\cite{faiz2024llmcarbonmodelingendtoendcarbon, wu2022sustainableaienvironmentalimplications, patterson2021carbonemissionslargeneural}. Operational carbon is computed as the product of E and CI, while embodied carbon is the product of Area and CPA, both outlined in Section~\ref{subsec:sustainable_computing}. For CI, we use world carbon intensity outlined in ACT~\cite{act_carbon}. We compute CPA with $E/mm^2$ detailed in Dark Silicon~\cite{8752110}, and convert it to CPA with the previously stated CI. 

\subsection{Simulation}
\label{sec:simulation}
We developed an in-house architecture simulator based on a publicly available artifact for Carat ~\cite{carat_paper}.
We extend the artifact and build an event-based simulator that can hierarchically solve the mapping of nonlinear operations and GEMM on our customized \name architecture and report area, leakage power, dynamic energy, cycle count, and runtime.
The basic hardware modules are implemented in RTL and synthesized under 400MHz with 45nm technology to retrieve metric values.
The memory access power are obtained from CACTI7~\cite{cacti7}.

We also place and route the full RTL of a single node 8x8 \name at 400MHz, and obtain area 0.056$mm^2$ and frequency of 408.5MHz with critical path on VLP broadcast.
We further increase the synthesis frequency of \name, and reach 975MHz without timing violations.
We stick to 400MHz to isolate of the impact of the implementation during the evaluation.


\section{Evaluation}
\label{sec:Evaluation}


\subsection{Nonlinear Approximation}
\label{subsec:nonlinear_approximation}
\subsubsection{Accuracy.}
\label{subsubsec:nonlinear_accuracy}
As shown in Figure~\ref{fig:nonlinear_heatmap}, \name softmax, SiLU, and GELU implementations achieve highly accurate results, outperforming other methods on most models.
Even in cases where Mugi is not the most accurate, Llama 2, it remains comparable to prior implementations.
Conversely, when value distributions are concentrated, Mugi yields substantial improvements in performance, as evident by the perplexity gains observed in Whisper Tiny and highlighted in Figure~\ref{fig:nonlinear_error}.
In contrast, prior approximation methods~\cite{c-lstm, mobilenet, exp_taylor} do not consider the value distribution of workloads prior to approximation, introducing additional error.


\begin{figure}[!t]
    \centering
    \includegraphics[width=\columnwidth]{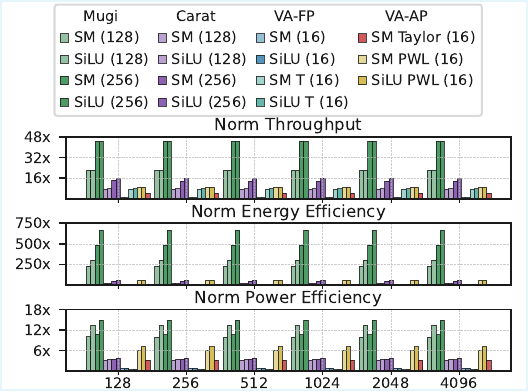}\\
    \Description{}
    \caption{Iso-area comparison between \name and baselines for nonlinear operations, across sequence lengths (128–4096) with a batch size of 8, geometric meaned across all Llama 2 models (7B, 13B, 70B).
    SM, and VA-FP, and VA-AP abbreviate softmax, precise vector array, and approximate vector arrays.
    Titles in the legend indicate the array type of each column.
    All results are normalized to VA (16) with array heights indicated in ``()''.}
    \label{fig:nonlinear_throughput}
\end{figure}

\subsubsection{Throughput and Efficiency.}
\label{subsubsec:nonlinear_performance}

Figure~\ref{fig:nonlinear_throughput} compares the throughput and efficiency. 
Since all designs map the head dimension across rows, sequence length does not impact the normalized performance gains.
\name achieves a shared normalized throughput of $45\times$, and energy and power efficiency improvements of $481.07\times$ and $10.69\times$ for softmax, and $667.85\times$ and $14.84\times$ for SiLU compared to precise vector arrays.
\name outperforms PWL approximation in softmax by $5\times$ (throughput), $8.53\times$ (energy efficiency), and $1.71\times$ (power efficiency), and in SiLU by $5\times$, $10.36\times$, and $2.37\times$, respectively.
Against Taylor series softmax, \name achieves $10.02\times$, $32.93\times$, and $3.28\times$ improvements in the same metrics.

\name contributes these gains in performance to its ability to scale to larger array sizes, sharing the compute array with GEMM. 
However, other designs require standalone vector arrays for nonlinear operations, where the scale is bounded by the SRAM bandwidth.
Additionally, \name does not have to compute costly exp~\cite{200964bitexpfunct}, thus multiplier free.
These improvements allow \name to increase throughput and efficiency compared to vector arrays for both precise and approximation configurations.



\minisection{Takeaway.}
\name enables accurate nonlinear approximation by applying input approximation and value-centric approach to VLP, greatly enhancing both throughput and efficiency.

\begin{figure}[!t]
    \centering
    \includegraphics[width=\columnwidth]{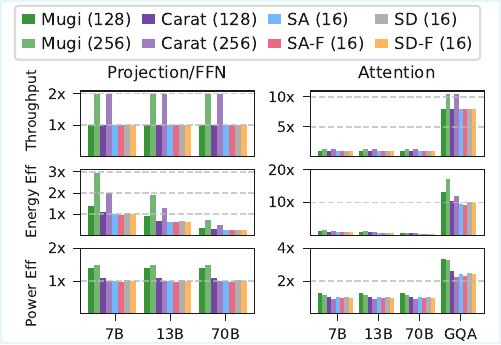}\\
    \Description{}
    \caption{Iso-area comparison for projection (proj), attention (attn), and feed-forward network (ffn) GEMM operations in Llama 2 (7B, 13B, 70B, and 70B with GQA).
    All results are normalized to that of 16$\times$16 systolic array (SA). 
    -F denotes FIGNA.
    Batch size is set to 8, and sequence length is set to 4096,
    with array heights indicated in parenthesis ().
    }
    \label{fig:gemm-breakdown}
\end{figure}

\begin{table}[!t]
    \centering
    \caption{Comparison of \name and baselines throughput, area, energy and power efficiency on LLaMa 2 70B (with GQA).
    Batch size is set to 8, and sequence length is set to 4096.
    Hardware details are outlined in Table~\ref{tab:hardware config}.}
    \begin{adjustbox}{max width=\linewidth}
    \begin{tabular}{c|c|c|c|c|c}
    \toprule
    \multicolumn{2}{c|}{\multirow{2}{*}{\textbf{Design}}} & \textbf{Throughput} & \textbf{OC Area} & \textbf{Energy Efficiency} & \textbf{Power Efficiency} \\
    \multicolumn{2}{c|}{} & \textbf{(Tokens/s)} & \textbf{($mm^2$)} & \textbf{(Tokens/s/$\mu$J)} & \textbf{(Tokens/s/W)} \\
    \midrule
    \multirow{8}{*}{SN} 
    & \name (128) & 0.71 & 2.16 & 68.64 & 3.18 \\
    & \name (256) & 1.39 & 3.10 & 142.82 & 3.37 \\
    & Carat (128) & 0.70 & 2.42 & 53.00 & 2.47 \\
    & Carat (256) & 1.38 & 3.84 & 95.78 & 2.27 \\
    & SA (16) & 0.67 & 2.58 & 45.97 & 2.24 \\
    & SA-F (16) & 0.67 & 2.81 & 44.34 & 2.16 \\
    & SD (16) & 0.67 & 2.54 & 47.83 & 2.34 \\
    & SD-F (16) & 0.67 & 2.77 & 46.06 & 2.25 \\
    \midrule
    \multirow{5}{*}{SN-S} 
    & SA (64) & 2.70 & 25.84 & 138.59 & 1.68 \\
    & SA-F (64) & 2.70 & 29.56 & 131.66 & 1.60 \\
    & SD (64) & 2.70 & 25.14 & 143.18 & 1.74 \\
    & SD-F (64) & 2.70 & 28.86 & 135.79 & 1.65 \\
    & Tensor & 10.06 & 38.75 & 488.31 & 1.59 \\
    \midrule
    \multirow{7}{*}{NoC}
    & 4 x 4 \name (256) & 22.19 & 50.12 & 2314.23 & 3.42 \\
    & 4 x 4 Carat (256) & 22.08 & 61.92 & 1551.09 & 2.30 \\
    & 4 x 4 SA (16) & 10.74 & 41.77 & 770.31 & 2.35 \\
    & 4 x 4 SA-F (16) & 10.74 & 45.48 & 741.68 & 2.26 \\
    & 4 x 4 SD (16) & 10.74 & 41.18 & 803.82 & 2.45 \\
    & 4 x 4 SD-F (16) & 10.74 & 44.89 & 772.70 & 2.36 \\
    & 2 x 1 Tensor (8) & 20.12 & 77.56 & 989.02 & 1.61 \\
    \bottomrule
\end{tabular}
    \end{adjustbox}
    \label{tab:design_comparison}
\end{table}


\begin{figure}[!t]
    \centering
    \includegraphics[width=\columnwidth]{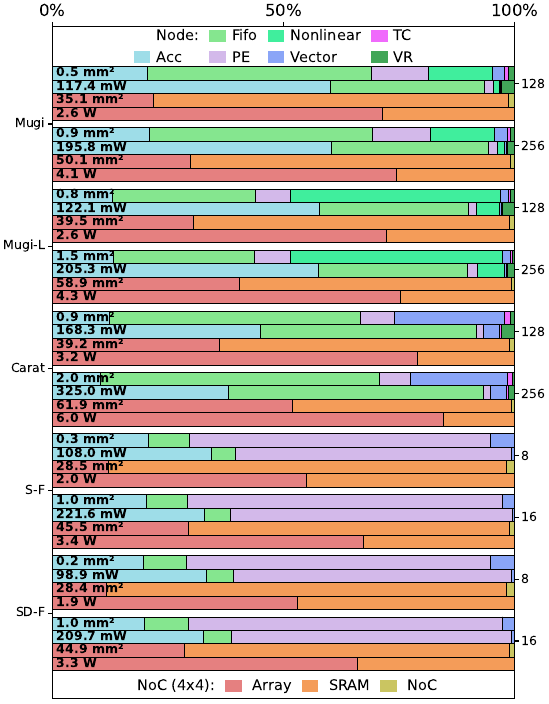}\\
    \Description{}
    \caption{Array and NoC level area and power breakdown.
        Total area and power is shown in each bar. Cool colored bars represent array level while warm colored bars represent NoC level.
        \textit{Acc} refers to output accumulators, and \textit{nonlinear} refers to nonlinear hardware. For the power breakdown, batch size is set to 8 and the sequence length is set to 4096.}
    \label{fig:area_power_breakdown}
\end{figure}



\subsection{GEMM} 
\label{subsec:gemm_op}
We show the GEMM execution results in Figure ~\ref{fig:gemm-breakdown}. 
The GEMM operations include the projection, attention, and FFN layers from the studied LLM models.
We observe that in terms of throughput and efficiency, \name consistently outperforms both systolic and SIMD arrays.
\name is optimized for 8 columns, which aligns with and benefits from a batch size and GQA group size of 8, allowing it to leverage GQA for better throughput.
This capability ensures that \name maintains excellent utilization even as the array size scales.
On the contrary, systolic and SIMD arrays face under-utilization with array sizes larger than 8x8.
Additionally, VLP eliminates multiplication in \name, increasing efficiency compared to baselines. 
Compared with Carat~\cite{carat_paper}, \name shows better efficiency as Carat needs additional hardware to execute nonlinear operations.

\minisection{Takeaway} \name achieves good throughput and efficiency gains through timely LLM optimizations in emerging asymmetric data formats, small-batch GEMM, and GQA.

\begin{figure*}[!t]
    \centering
    \includegraphics[width=\textwidth]{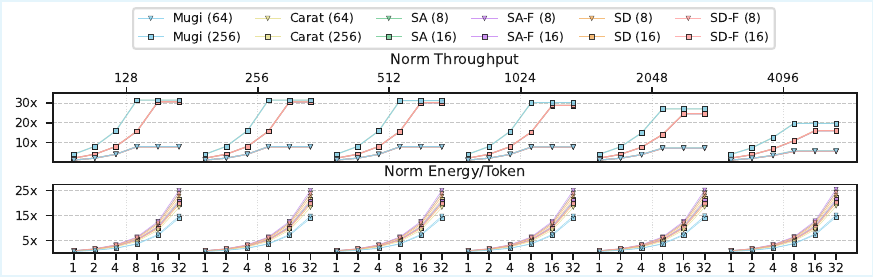}
    \caption{ISO-throughput LLMs study, geometric mean over all Llama models.
    The x-axis represents batch size at the bottom and each plot represents sequence lengths (128-4096) at the top.
    The y-axis represents the improvements of throughput and energy per token.
    All designs are normalized to an 8$\times$8 systolic array with batch a size of 1. 
    -F denotes FIGNA, and array heights are indicated in ``()''.
    SA and SD base and -F array's throughput closely overlap, as do \name and Carat.}
    \label{fig:batch_size_impact}
\end{figure*}


\subsection{LLM workload}
\subsubsection{Single Node}
\label{subsubsec:single_node_evalution}

Single-node evaluations show that \name exceeds baseline implementations in both throughput and efficiency, while also reducing overhead and area costs. An end-to-end comparison is provided in Table~\ref{tab:design_comparison}. Compared to a baseline systolic array 16, the \name 256 architecture achieves an increase of $2.07\times$, $3.11\times$, and $1.50\times$ in throughput, energy efficiency, and power efficiency respectively.
These improvements stem from \name’s effective reuse of the compute array for both nonlinear operations and GEMM, and efficient handling of asymmetric, small-batch GEMM using WOQ, KVQ, and GQA.


Figure~\ref{fig:area_power_breakdown} highlights these advantages, which are further amplified by the elimination of specialized vector arrays and costly MAC units, resulting in a more compact compute array and lower power consumption. 
\name additionally scales more efficiently to larger array sizes, growing linearly.
On the contrary, the area of systolic and SIMD arrays scales up quadratically as the array scales in both row and column dimensions.
Though also based on VLP, Carat area scales up super-linearly, due to the excessive cost of FIFO.
Despite the large area, the temporal coding in VLP still minimizes the switching activities and ensures a low power-to-area ratio.
Comparing to \name, \name-L with LUT for nonlinear operations and VLP for GEMM, spends way more hardware on on-chip LUT, implemented using FIFOs to ensure programmability.

We further show an extended throughput and energy comparison of different designs when sweeping the batch size, as shown in Figure~\ref{fig:batch_size_impact}, offering higher throughput and lower energy per token.
The best throughput of \name is attainable at a smaller batch size of 8 than other baselines, as \name maps the batch across columns and peaks the utilization at a batch size of 8.

We additionally considered off-chip memory accesses, and we see that \name handles DRAM traffic similarly to other dataflow architectures, almost identical operational intensity, but offers higher compute utilization, therefore more compute bounded.

Figure~\ref{fig:latency_breakdown} shows a latency breakdown of different LLMs.
It is observed that \name have slightly better latency for attention GEMMs than other designs, but almost halves the latency for projection and FFN GEMMs.
Regarding the nonlinear operations, \name shows almost invisible latency, exhibiting tremendous improvements over other designs.
Though not obvious, Carat triples the nonlinear latency of \name, due to relying on non-VLP approximations.

\subsubsection{Carbon Emission}
\label{subsubsec:carbon_eval}
\begin{figure}[!t]
    \centering
    \includegraphics[width=\columnwidth]{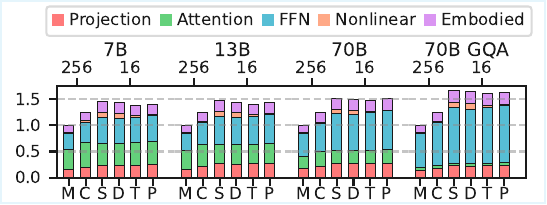}\\
    \caption{Normalized onchip operational and embodied carbon across model sizes (Llama 2-7B, 13B, 70B, 70B GQA) of \name and baseline configurations. Batch size is set to 8, and sequence length is set to 4096. Top x-axis represents array height, while y axis is normalized latency. M, C, S, D, T, P represents \name, Carat, Systolic, SIMD, Taylor Series, and Piecewise linear approximate respectively.}
    \label{fig:carbon_breakown}
\end{figure}

\begin{figure}[!t]
    \centering
    \includegraphics[width=\columnwidth]{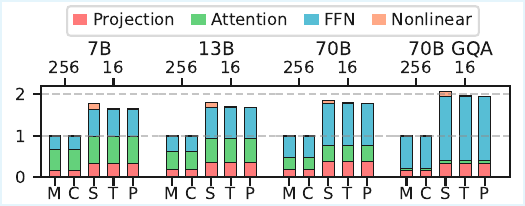}\\
    \caption{Normalized end-to-end latency breakdown across model sizes.
    All notations are identical to those in Figure~\ref{fig:carbon_breakown}, except that S here is for Systolic/SIMD.
    }
    \label{fig:latency_breakdown}
\end{figure}

When comparing \name's carbon impact to baselines, we see that \name improves in both operational and embodied emissions by $1.45\times$ and $1.48\times$ respectively.
While Figure~\ref{fig:carbon_breakown} shows operational carbon as the major contributor to emissions, this follows previous trends consitent with 45nm technologies.
Through \name's efficient, shared compute array, \name is able to simultaneously decrease both operational and embodied carbon while delivering an increase to throughput for LLM workloads.

\begin{figure}[!t]
    \centering
    \includegraphics[width=\columnwidth]{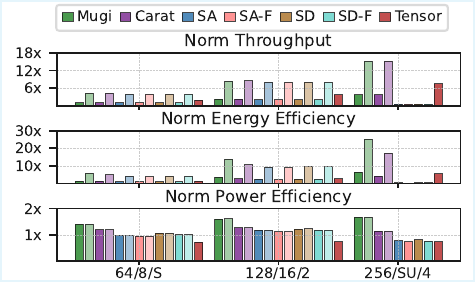}\\
    \caption{Normalized NoC-level throughput, energy efficiency, and power efficiency for \name and baseline 4x4 and 8x8 NoC architectures.
    Data represents the geometric mean across model sizes (Llama 2–7B, 13B, 70B) with a batch size of 8 and sequence length set to 4096.
    The x-axis represents array height for each group, in order of VLP, SA/SD, and tensor core.
    Tensor core configurations represent -S (single node), 2 (2x1 NoC), 4 (2x2 NoC). All models are normalized to an 8$\times$8 systolic array (NoC dim: 4x4).}
    \label{fig:noc}
\end{figure}

\subsubsection{Multi Node}
\label{sec:multi_node_mugi}
\name scales efficiently to multi-node designs using a NoC architecture. 
We compare \name’s multi-node implementations against baseline designs with the same NoC configuration and scaled-out versions of single node baseline architectures.
As with single node setups, \name’s multi node configurations show comparable gains in throughput, energy efficiency, and power efficiency when comparing \name 4x4 256 and systolic array 4x4 16.
Figure~\ref{fig:noc} details these improvements, emphasizing the benefits of multi-node architectures.
Moreover, NoC-based implementations clearly outperform scaled-up systolic arrays, due to severe under-utilization at a small batch size, as shown in Table~\ref{tab:design_comparison}. 
Figure~\ref{fig:area_power_breakdown} shows the breakdown of NoC level area, and the array is occupying varying ratios of the on-chip resource, given an identical on-chip SRAM size.

\minisection{Takeaway}
By enabling VLP for both nonlinear operations and GEMM, \name effectively accelerates all aspects of LLM workloads, which also scales to multi nodes.

\section{Discussion}
\label{sec:Discussion}

\subsection{Limitations}
\label{subsec:limitations}
While we showcase the benefits of VLP, there are a few workloads or techniques not addressed in this paper.

\minisection{Additional Operations.}
While \name unlocks nonlinear approximation, there are still some nonlinear LLM operations not supported, such as layer normalization and rotary positional embeddings (RoPE).
Layer normalization are vector multiplication, and can be supported in \name's vector unit.
For RoPE, \name can either approximate the required sine and cosine functions, though the utilization might be low due to its sparse nature, or offload them to external hardware as in existing GEMM accelerators.

\minisection{MoE and Multi-Modal Models.}
Mixture-of-Experts (MoE) extend standard attention-based LLMs with selective FFN experts, selected by a softmax-based gating network~\cite{GLaM, gshard, deepseekmoe}.
Multi-modal models support multiple modalities beyond just text (i.e., language, vision, video, etc), by either tokenizing non-language inputs~\cite{palm-e}, or combining multiple modality-specific layers~\cite{qwen3vl, gemini_mm}.
There additionally exist models that leverage MoE architecture on multiple modalities~\cite{meta_llama4}.
All these additional operations have been supported in \name, and different modality has been studied in this work to prove the efficacy of VLP.
We conjecture \name should generalize to both MoE and multi-modal variants, though we leave full validation to future work.

\minisection{Online Approximation.}
Currently, \name pre-computes the results offline for accurate nonlinear approximation via LUT.
However, the value distribution could exhibit a slight drift at runtime.
Such drifts in both KV cache and FFN have been well tackled via quantization, using as few as 4 bits.
As for softmax, since all softmax inputs are subtracted by the maximum for numerical stability, the drift minimally impacts accuracy.
Moreover, \name’s sliding window mechanism helps adapt to the current workload and further reduces the impact of drift.
That said, we argue optimal accuracy would benefit from an online mechanism to adjust LUT values at runtime, and we leave this to future work.



\subsection{Related Works}
\label{subsec:related_works}

\minisection{Nonlinear Approximation.}
Prior works have explored approximating nonlinear operations.
Some approaches use piece-wise linear (PWL) approximations, partitioning the function into segments and selecting coefficients based on input range, while others approximate the entire function with a single simplified equation~\cite{mobilenet, c-lstm, 2019dac_instream, 2021dt_instream}.
Other works use Taylor-series expansions, which can provide high accuracy but degrade as values drift from the expansion point~\cite{exp_taylor, spatten}.
While all approximation techniques, including \name, introduce some levels of approximation error, others underperform \name in most models while increasing area and efficiency costs, as detailed in Figure~\ref{fig:nonlinear_heatmap} and Table~\ref{tab:design_comparison}.

\minisection{GEMM Acceleration.}
A number of prior works target GEMM acceleration.
Carat first introduced VLP, enabling low-precision, large-batch value reuse for CNNs~\cite{carat_paper}.
While large batch sizes compliment CNN workloads, LLMs operate with smaller batch sizes and larger matrices, making Carat unsuited for such workloads.
Other accelerators exploit unary computing~\cite{2020isca_ugemm, 2021microtoppick_ugemm, 2022hpca_usystolic, 2022isca_ubrain} to reduce buffer overhead using ternary RIM arrays and unary half-adders~\cite{cambricon}, improving accumulation efficiency.
However, each PE still requires multipliers and accumulators, whereas \name eliminates costly PE MAC units altogether via VLP.
Additional works identify the ability to exploit data reuse to reduce data movement.
Multi-chip module designs share inputs across weight-stationary chiplets to exploit cross-chiplet reuse~\cite{Simba}.
Matrix-scaling approaches reduce large matrices into sub-matrices and scaling vectors, reducing memory transfers sharing sub-matrix computation during rescaling~\cite{mecla}.
Lastly, another work identifies that quantization reduces the number of possible inputs, enabling computation to be shared between layers where outputs do not change~\cite{computation_reuse}.
Like \name, these works similarly identify data reuse techniques but largely overlook value reuse, missing further opportunities for optimization.

\minisection{KV Cache Compatibility.}
Some works employ detailed hardware–software co-design to improve LLM performance.
These works aim to reduce computation through top-k speculative prediction, removing computation predicted to be negligible to the attention output and reducing KV cache accesses~\cite{mcbp, elsa, fact, spatten}.
In contrast, other accelerators focus purely on GEMM computation and do not address attention or KV-cache related bottlenecks~\cite{carat_paper, cambricon, Simba, computation_reuse, mecla}.
\name occupies a middle ground by incorporating lightweight KV-cache optimizations natively in its architecture, avoiding workload-specific optimizations via co-design.
This allows \name to generalize across models while still capturing key reuse opportunities in modern LLM workloads.

\section{Conclusion}
\label{sec:Conclusion}

In this paper, we orchestrate value-level parallelism (VLP) for efficient transformer-based LLMs.
We formulate VLP for nonlinear approximation in a value-centric approach where important values are assigned with greater accuracy.
We design a \name architecture for our VLP approximation.
Additionally, we optimize \name to accelerate asymmetric, small-batch GEMM, which leverages the trending LLM optimizations.
To this end, \name effectively supports full LLM workloads via VLP.
Our experimental results demonstrate significant performance and efficiency gains in \name, highlighting the potential of VLP for AI workloads.









\clearpage
\appendix
\section{Artifact Appendix}

\subsection{Abstract}

The artifact evaluation is separated into two scopes, workload evaluation and architecture evaluation.
The workload evaluation contains nearly all figures within section~\ref{sec:Mapping}, namely Figure~\ref{fig:nonlinear_distribution}, Figure~\ref{fig:nonlinear_heatmap}, and Figure~\ref{fig:nonlinear_error}.
The architecture evaluation includes all figures within section~\ref{sec:Evaluation}.
Both artifacts can be run on an x86\_64 machine with python and conda installed.
The workload evaluation requires access to a GPU cluster.
We have tested the architecture evaluation on ubuntu 24.04, and the workload evaluation on an NSF GPU cluster running Red Hat 9.4.
To run each artifact, download the artifacts from the zenodo links detailed in Section~\ref{subsec:artifact_description} and follow steps outlined in Section~\ref{sec:Artifact-install}.
To see the results generated from running the artifact, see Section~\ref{sec:Artifact-results} for detail.

\subsection{Artifact check-list (meta-information)}

\begin{itemize}
  \item {\bf Workload evaluation:}
    \begin{itemize}
      \item {\bf Model:} AI profiling
      \item {\bf Data set:} Models outlined in Table~\ref{tab:studied_llms}.
      Note that Llama 70B results are excluded in the provided artifact, due to the excessive profiling time.
      \item {\bf Run-time environment:} Conda
      \item {\bf Hardware:} NSF GPU cluster
      \item {\bf Metrics:} Perplexity, value distribution, theoretical error
      \item {\bf Output:}
        Figure~\ref{fig:nonlinear_heatmap}, Figure~\ref{fig:nonlinear_distribution}, Figure~\ref{fig:nonlinear_error},
      \item {\bf Experiments:} Approximate perplexity comparison, model value distribution, approximation theoretical error
      \item {\bf Disk space required (approximate):} 70GB
      \item {\bf Time needed to complete experiments (approximate):} 12-24 hours
      \item {\bf Publicly available:} Yes
      \item {\bf Code licenses (if publicly available):} MIT License
      \item {\bf Workflow framework used:} Pytorch
    \end{itemize}

  \item {\bf Architecture evaluation:}
    \begin{itemize}
        \item {\bf Model:} Cycle-level performance model, event-based cost model
      \item {\bf Data set:} Llama configurations outlined in Table~\ref{tab:studied_llms}
      \item {\bf Run-time environment:} Conda
      \item {\bf Hardware:} x86\_64 machine
      \item {\bf Metrics:} Throughput, latency, energy efficiency, power efficiency, area, carbon equivalent emissions.
      \item {\bf Output:}
        Figure~\ref{fig:nonlinear_throughput}, Figure~\ref{fig:gemm-breakdown}, Table~\ref{tab:design_comparison}, Figure~\ref{fig:area_power_breakdown}, Figure~\ref{fig:batch_size_impact}, Figure~\ref{fig:carbon_breakown}, Figure~\ref{fig:latency_breakdown}, Figure~\ref{fig:noc}.
      \item {\bf Experiments:} Iso-area nonlinear comparison, iso-area GEMM comparison, comprehensive design comparison,  array and noc-level area and power breakdown, batch size comparison, operational and embodied carbon comparison, end-to-end latency comparison, iso-area noc comparison.
      \item {\bf Disk space required (approximate):} 8GB
      \item {\bf Time needed to complete experiments (approximate):} 0.5 - 1 hours
      \item {\bf Publicly available:} Yes
      \item {\bf Code licenses (if publicly available):} MIT License
      \item {\bf Workflow framework used:} In-house simulation framework
    \end{itemize}
\end{itemize}

\subsection{Description}
\label{subsec:artifact_description}

\subsubsection{How to access}
\label{subsec:how_to_access}
Both artifacts can be downloaded at \textcolor{blue}{\texttt{\url{https://zenodo.org/records/18063514}}}.
Follow the instructions detailed in the zenodo or Section~\ref{sec:Artifact-install} and \ref{sec:Artifact-workflow} to evaluate each artifact.
\subsubsection{Hardware dependencies}
For the workload evaluation, a GPU capable of loading all models at half precision is required.
\subsubsection{Software dependencies}
Conda is required to build the environment.\footnote{Available at https://www.anaconda.com/download.}
\subsubsection{Datasets}
Both evaluations use the models outlined in Table~\ref{tab:studied_llms}, with the architecture evaluation only including Llama models, and the workload evaluation ignoring Llama 2 70b.
Access to the ML models is provided in the artifact during the artifact evaluation, but will be deprecated after evaluation, and access to the models will have to be obtained.\footnote{Available at https://huggingface.co/meta-llama.}
\subsubsection{Models}
The models used in our simulation framework include a cycle-level performance model and an event-based cost model, as well as profiling and end-to-end perplexity results.
\subsection{Installation}
\label{sec:Artifact-install}
One may follow the steps below to run the artifact, also available at the zenodo link in Section~\ref{subsec:how_to_access}.
For the artifact evaluation, we will provide access to the NSF GPU cluster.
Both evaluations detail the steps assuming a Linux environment, and require the command-line tool unzip to be installed prior to evaluation.
If you are evaluating either artifact on a local machine, follow the commands below to install unzip.
Otherwise, ensure that unzip or and equivalent command-line tool is available.

\textcolor{blue}{\texttt{sudo apt update}}

\textcolor{blue}{\texttt{sudo apt install unzip}}

\minisection{Workload evaluation.}
\begin{enumerate}
    \item Download the zip file from the zenodo link.

    \textcolor{blue}{\texttt{mugi\_profiling-asplos\_2026\_ae.zip}}
    
    \textcolor{blue}{\texttt{\url{https://zenodo.org/records/18063514}}}
    
    


    \item Unzip the artifact and cd into the new directory.

    \textcolor{blue}{\texttt{unzip mugi\_profiling-asplos\_2026\_ae.zip}}

    \textcolor{blue}{\texttt{cd mugi\_profiling-asplos\_2026\_ae}}

    \item Create a conda environment with the included environment.yaml and activate the environment.

    \textcolor{blue}{\texttt{conda env create -f environment.yaml}}
    
    \textcolor{blue}{\texttt{conda activate mugi\_profiling}}
    \item Run the included script to launch all slurm scripts.

    \textcolor{blue}{\texttt{bash mugi\_profiling.sh}}
    
    If slurm access is not available (NSF access fails), an additional script is provided, but tuning may be required.
    
    \textcolor{blue}{\texttt{bash mugi\_profiling\_local.sh}}

    \item Finally, retrieve the figures in the figures/output directory.
\end{enumerate}

\minisection{Architecture evaluation.}
\begin{enumerate}
    \item Download the zip file from the zenodo link.

    \textcolor{blue}{\texttt{archx-asplos\_2026\_ae.zip}}
    
    \textcolor{blue}{\texttt{\url{https://zenodo.org/records/18063514}}}
    

    
    

    \item Unzip the artifact and cd into the new directory.

    \textcolor{blue}{\texttt{unzip archx-asplos\_2026\_ae.zip}}

    \textcolor{blue}{\texttt{cd archx-asplos\_2026\_ae}}

    \item Create a conda environment with the included environment.yaml and activate the environment.

    \textcolor{blue}{\texttt{conda env create -f environment.yaml}}

    \textcolor{blue}{\texttt{conda activate archx}}
    \item Run the included script to generate and simulate the architecture descriptions.

    \textcolor{blue}{\texttt{bash run\_mugi.sh}}
    \item Finally, retrieve the figures in the zoo/llm/results/figs/ and zoo/llm/results/tables/ directories.
\end{enumerate}

\subsection{Experiment workflow}
\label{sec:Artifact-workflow}
\minisection{Workload evaluation.}
We use slurm scripts to automate profiling and end-to-end perplexity results.
To produce each figure, the scripts load the model onto an allocated node and run the target dataset.
Profiling is done on the base half-precision models, while perplexity is retrieved for base models at both half precision and for nonlinear approximation.
After all models run, the scripts process both the profiling and perplexity results and report them within each figure.

\minisection{Architecture evaluation.}
We use scripts to automatically run the workflow for the result production.
To produce a figure, the evaluation scripts first generate all the hardware configurations based on the provided architecture descriptions.
Then, the simulation is run on each architecture description with a target workload (e.g., llama).
More specifically, both the performance model and cost model are run.
Note that the automated workflow launches these simulations in parallel.
Finally, the scripts parse the generated result and aggregate results across runs from all the configurations to generate figures.
\subsection{Evaluation and expected results}
\label{sec:Artifact-results}
After running the steps in Section~\ref{sec:Artifact-install}, the generated figures can be found locally in \verb|figures/output| directory for workload evaluation and \verb|zoo/llm/results/figs| and \\
\verb|zoo/llm/results/tables| for the architecture evaluation.
Each figure is labeled as \verb|figX-Y.pdf| that corresponds to what is included in Section~\ref{sec:Mapping} and Section~\ref{sec:Evaluation}.
The expected profiling distribution in the workload evaluation may exhibit slight deviations from the values reported in the paper due to device-specific computational variation.



\subsection{Methodology}

Submission, reviewing and badging methodology:

\begin{itemize}
  \item \url{https://www.acm.org/publications/policies/artifact-review-badging}
  \item \url{http://cTuning.org/ae/submission-20201122.html}
  \item \url{http://cTuning.org/ae/reviewing-20201122.html}
\end{itemize}

\clearpage

\bibliographystyle{ACM-Reference-Format}
\bibliography{ref}

\end{document}